\documentclass[conference]{IEEEtran}
\IEEEoverridecommandlockouts
\usepackage{cite}
\usepackage{amsmath,amssymb,amsfonts}
\usepackage{graphicx}
\usepackage{booktabs}
\usepackage{textcomp}
\usepackage{xcolor}
\usepackage[ruled,vlined,linesnumbered]{algorithm2e}
\usepackage{amsmath}
\usepackage{amssymb}
\usepackage{pifont}
\newcommand{\cmark}{\ding{51}}%
\newcommand{\xmark}{\ding{55}}%
\newcommand{\olsi}[1]{\,\overline{\!{#1}}} %
\def\BibTeX{{\rm B\kern-.05em{\sc i\kern-.025em b}\kern-.08em
    T\kern-.1667em\lower.7ex\hbox{E}\kern-.125emX}}
\usepackage{hyperref}
\hypersetup{
    colorlinks=true,
    linkcolor=blue,
    filecolor=blue,      
    urlcolor=blue,
    pdftitle={JCET2022},
    pdfpagemode=FullScreen,
}
\begin{document}

\title{Quantifying Causes of Arctic Amplification via Deep Learning based Time-series Causal Inference}

\author{
\IEEEauthorblockN{
Sahara Ali\IEEEauthorrefmark{1}\IEEEauthorrefmark{4},
Omar Faruque\IEEEauthorrefmark{1},
Yiyi Huang\IEEEauthorrefmark{1},
Md. Osman Gani\IEEEauthorrefmark{1}\IEEEauthorrefmark{4},
Aneesh Subramanian\IEEEauthorrefmark{2}\IEEEauthorrefmark{4},
Nicole-Jeanne Schlegel\IEEEauthorrefmark{3},
\\ Jianwu Wang\IEEEauthorrefmark{1}\IEEEauthorrefmark{4}
}

\IEEEauthorblockA{
\IEEEauthorrefmark{1}Department of Information Systems,
      University of Maryland, Baltimore County, Baltimore, MD, United States
\IEEEauthorblockA{
\IEEEauthorrefmark{2}Department of Atmospheric and Oceanic Sciences, University of Colorado Boulder,                   Boulder, CO, United States}
\IEEEauthorblockA{
\IEEEauthorrefmark{3}Geophysical Fluid Dynamics Laboratory, NOAA, Princeton, NJ, United States}
\IEEEauthorblockA{
\IEEEauthorrefmark{4}NSF HDR Institute for Harnessing Data and Model Revolution in the Polar Regions (iHARP), United States}
\IEEEauthorblockA{ Emails: \IEEEauthorrefmark{1}\{sali9,omarf1,yhuang10,mogani,jianwu\}@umbc.edu, \IEEEauthorrefmark{2}aneeshcs@colorado.edu, \IEEEauthorrefmark{3}nicole.schlegel@noaa.gov
}
}
}

\maketitle

\begin{abstract}
  The warming of the Arctic, also known as Arctic amplification, is led by several atmospheric and oceanic drivers. However, the details of its underlying thermodynamic causes are still unknown. Inferring the causal effects of atmospheric processes on sea ice melt using fixed treatment effect strategies leads to unrealistic counterfactual estimations. Such methods are also prone to bias due to time-varying confoundedness. Further, the complex non-linearity in Earth science data makes it infeasible to perform causal inference using existing marginal structural techniques. In order to tackle these challenges, we propose TCINet - \textbf{T}ime-series \textbf{C}ausal \textbf{I}nference \textbf{Net}work to infer causation under continuous treatment using recurrent neural networks and a novel probabilistic balancing technique. More specifically, we propose a neural network based potential outcome model using the long-short-term-memory (LSTM) layers for time-delayed factual and counterfactual predictions with a custom weighted loss. To tackle the confounding bias, we experiment with multiple balancing strategies, namely TCINet with the inverse probability weighting (IPTW), TCINet with stabilized weights using Gaussian Mixture Model (GMMs) and TCINet without any balancing technique. Through experiments on synthetic and observational data, we show how our research can substantially improve the ability to quantify leading causes of Arctic sea ice melt, further paving paths for causal inference in observational Earth science.

\end{abstract}


\section{Introduction}\label{sec:intro}
In the last few decades, Earth and Atmospheric scientists have observed greater climate change near the polar regions as compared to the rest of the world \cite{lee2014theory}. In 2018, the observed mean sea ice extent (SIE) at Kara Sea during the summer months of June, July and August (JJA) reduced to half of what it was in 1979, i.e. from 1.25 $million$ $km^2$ to just 0.5 $million$ $km^2$. What we are observing can happen in response to a change in global climate forcing. Due to the melting of highly reflective sea ice and snow regions in the Arctic and Antarctic, there is an increased absorption of solar radiation which amplifies the warming. This phenomenon, also known as polar amplification is causing the melting of polar ice sheets,  resulting in sea level rise, and the rate of carbon uptake in the polar regions \cite{lee2014theory}. In light of this phenomenon, the warming of Arctic sea ice is referred to \textbf{Arctic Amplification} \cite{rantanen2022arctic}. Though, it has not been scientifically proven if the Arctic has warmed more than the rest of the hemisphere, studying the cause of thinning and retreat of the Arctic sea ice is a significant and substantial topic of atmospheric research \cite{holland2003polar}.

In this paper, we dive deeper into the concept of time-series Causal Inference (CI) and present the challenges and opportunities for performing CI to study Arctic amplification under continuous treatment effect. Causal inference can be defined as the process of estimating the causal effects (influence) of one event, process, state or object (a cause) on another event, process, state or object (an effect). For estimation of causal effect, there are two main categories of techniques, potential outcome framework and do-calculus. The potential outcome framework relies on hypothetical interventions such that it defines the causal effect as the difference between the outcomes that would be observed with and without exposure to the intervention \cite{rubin1990comment}. This technique is widely used in epidemiology where patients are randomly divided into treated and controlled groups and the effectiveness of treatment is inferred by observing patients condition with and without undergoing a treatment \cite{rubin2005causal}. The treatment effect can be measured at individual, treated group, sub-treated group and entire population levels \cite{moraffah2021causal}. At the population level, the treatment effect measured is called Average Treatment Effect (ATE). In case of Earth science observational data, we consider ATE a more suitable metric as a causal estimand which quantifies the mean difference observed in potential outcomes $Y$ given the exposure to treatment $X$, i.e., $Y(X=1)$ versus the inexposure to the treatment, i.e., $Y(X=0)$.
The established standard approach of performing time-varying causal inference in case of linear time-series data is through the use of marginal structural models \cite{robins2000marginal}, whereas, recent development in deep learning has paved paths for robust techniques for performing causal inference on non-linear observational and longitudinal data \cite{koch2021deep}. While existing deep learning models majorly handle independent and identically distributed (i.i.d) data \cite{koch2021deep}, we see only a handful of techniques capable of joint representation learning of continuous treatment and covariates in time-varying setting \cite{lim2018forecasting, bica2020estimating, bica2020time}. We compare in Table \ref{tab:tsci-methods} the capability of existing deep learning and machine learning methods in fulfilling Earth science requirements.

In light of above background, this paper presents a deep learning based time-series causal inference method that overcomes the limitations of existing causal effect estimation approaches in answering important research questions pertaining to the climate change effects in the Arctic. We present TCINet, a deep learning based time-series causal inference model, for counterfactual prediction under time-delayed continuous treatment. Our major contributions can be summarized as follows: 
\begin{itemize} 
    \item We propose a deep learning based time-series causal inference model suitable for both time-varying and time-invariant treatment effect estimation, which includes a new definition for average treatment effect estimation in case of time-delayed continuous treatment.
    \item We propose a novel probabilistic weighting technique to balance time-varying confoundedness by leveraging Gaussian Mixture Model (GMM).
    \item We perform extensive experiments evaluating our approach and compare it with the state-of-the-art (SOTA) approaches using synthetic time series data for fixed and continuous time-delayed treatments; further verifying our quantified causal effect results of thermodynamic processes on the Arctic sea ice melt with domain knowledge.
\end{itemize}

Moving forward, we will use the following terminologies: treatment variable (which is an identified cause), potential outcome (the variable identified as effect) and covariates (a set of variables that are either common cause of both treatment and outcome or are descendants of treatment variables identified in the causal graph).

\subsection{Formulating Causal Inference for Earth Science}
Climate data is non-stationary with climatological trends and visible annual seasonality cycles, therefore, binary or fixed treatment effect estimation can be an unrealistic way of quantifying causation. Further, in the absence of ground truth, the exposure to a policy change or applying dynamic treatment regime cannot be observed. This leads to the inability to evaluate model's performance for counterfactual predictions in observational data \cite{stuart2004matching, runge2019inferring}. Existing techniques such as marginal structural models \cite{robins2000marginal}, time-series regression \cite{google45950}, matching methods \cite{stuart2004matching} and deep learning based counterfactual predictions struggle in accurately inferring causation under time-delayed continuous intervention \cite{koch2021deep}. To fill this gap, we propose a deep learning based inference model, based on the potential outcomes framework \cite{rubin1990comment} and extending the recurrent methods based counterfactual approach \cite{lim2018forecasting} to study the impact of time-delayed treatment in the presence of time-varying covariates.

More formally, given treatment $X_t$ at timestep $t$, our model infers the time-delayed outcome $Y_{t+l}$ at $l$ steps ahead in future, in the presence of a set of $M$ time-dependent covariates $Z_{t}$. We give a generic formulation of our problem as follows. $Y(\hat{X_t})$ is the potential outcome, i.e., forecasted values under intervened treatment $\hat{X}$ at time $t$, and $Y(X_t)$ is the potential outcome under treatment $X$ at time $t$ without intervention (also called placebo effect), whereas $Z_{t}$ represents the covariates at time $t$, and $f$ represents our proposed deep learning based inference model. We utilize both factual and counterfactual predictions of $Y$ for all $N$ timesteps to estimate \textbf{\textit{lagged average treatment effect (LATE)}} under continuous intervention:  
\begin{equation}
   Y_{t+l}(X=x_t) = f(Z_{t},x_t) \\
\end{equation}
\begin{equation}
   Y_{t+l}(\hat{X}=\hat{x}_t) = f(Z_{t},\hat{x_t}) \\
\end{equation}
\begin{equation}
   LATE(l) = \frac{1}{N} \sum_{t=1}^{N} E [Y_{t+l}(\hat{X_{t}}) - Y_{t+l}(X_{t})]
   \label{eq:late}
\end{equation}

For consistent causal effect estimation under time-varying treatment, our proposed model holds the standard identifiability conditions or causal assumptions of consistency, positivity and conditional exchangeability \cite{robins2000marginal, moraffah2021causal}. Our implementation code can be accessed at the iHARP GitHub repository\footnote{ \href{https://github.com/iharp-institute/causality-for-arctic-amplification/tree/main/tcinet-icmla2023}{github.com/iharp-institute/causality-for-arctic-amplification}}.



\section{Related Work}
Though causality based study is a comparatively a new paradigm in Earth science, causal inference has been a widely studied topic for decades in statistics, economics, public policy and even healthcare \cite{yao2021survey, moraffah2021causal, koch2021deep}.

\subsubsection{G-Methods for Time-Varying Causal Inference}
Estimating time-varying causal or treatment effects leads to the problem of time-varying confounding, that is the common influence a past treatment or covariate might have on the future treatments and the future outcome. Robin's g-methods have shown to provide promising results on reducing bias caused by time-varying treatment and covariates on the potential outcome \cite{naimi2017introduction}. G-methods provide metrics to overcome the problem of time-varying confounding through standardization, g-computation, and inverse probability of treatment weighted (IPTW) estimators \cite{naimi2017introduction}. The prediction models of these estimators are typically based on linear or logistic regression such as Causal-ARIMA \cite{menchetti2021estimating}, Time Based Regression (TBR)\cite{google45950} and Marginal Structural Models (MSMs)\cite{robins2000marginal}. One big limitation of these methods is that, in case of complex non-linearity in treatment or outcome variables, the methods will lead to inaccurate results.  

\subsubsection{Deep Learning based Causal Inference}
Causal inference methods based on representation learning or deep learning techniques \cite{bengio2013representation} learn the representation of input data by extracting features from the covariate space \cite{koch2021deep}, where majority of the existing deep learning based methods are developed for i.i.d data \cite{koch2021deep}. In these deep learning based CI methods, a single neural network (also called meta learner) can be trained to make predictions for both treatment and control groups individually to estimate the average treatment effect (ATE). Existing meta-learners include S(ingle)-learner \cite{kunzel2019metalearners}, and T-learner or multi-task learners \cite{johansson2022generalization, shalit2017estimating} that jointly predict outcome for treated and controlled groups. X-learner \cite{kunzel2019metalearners} or cross-group learners are a hybrid form of meta learners that overcome the problem of unbalanced data in treatment and controlled groups. U-learner \cite{nie2021quasi} and R-learner utilizes Robinson transformation to develop a custom loss function for conditional ATE estimation \cite{nie2021quasi}. SCIGAN is another causal inference method for estimating the effects of continuous-valued interventions that aim to learn the distribution of unobserved counterfactuals using Generative Adversarial Networks (GANs) \cite{bica2020gan}. The limitations of CI methods for i.i.d. data is that these methods perform poorly on sequential or time-series data with no capability to handle time lags or time-varying confounding effects, thereby leading to invalid causal effect estimation results. 
For time-series causal inference, researchers have proposed methodologies based on machine learning and deep learning models that can also tackle the problem of time-varying confounding \cite{moraffah2021causal}. Recurrent Marginal Structural Networks (R-MSN) \cite{lim2018forecasting} and Counterfactual Recurrent Network (CRN) \cite{bica2020estimating} are some of the recent models that claim to estimate causal effects in the presence of time-varying confounders, however, contrary to the claim, these methods are healthcare-specific and cannot be generalized for other domain areas like Earth science because these models require on-hot encoded treatment flags with multivariate combined dosage. Talking about counterfactuals, the most recent model, Time Series Deconfounder - a multi-task method, leverages the assignment of multiple treatments over time to enable the estimation of treatment effects in the presence of multi-cause hidden confounders \cite{bica2020time}. The Conditional Instrumental Variable (CIV) \cite{CIV-2022} method measures the causal effect $\beta$ from covariates to the target variable using instrument variables that have a relation with covariates and target but are independent of any hidden confounder. To yield a better estimation the instrument variables are conditioned for single or multiple previous time steps in CIV.
Though deep representation learning methods are capable of automatically learning the intrinsic correlations and are also effective in accurate counterfactual estimation, they often lead to predictions with high variance or uncertainty estimates. 

\subsubsection{Time-Varying Causal Inference for Earth Science}
From climate or atmospheric science perspective, causality remains a lesser tapped area \cite{jerzak2023integrating, runge2019inferring, runge2023causal} and climatologist still rely on dynamical modeling techniques where certain atmospheric variables are nudged or perturbed as initial conditions in the physical simulation models (also called Earth System Models) to evaluate the outcome of these interventions on target variables \cite{Marcovecchio2021, van2021methodology, huang2019survey}.   
Applying deep learning techniques to infer causal effects of climate change offers a data-driven and cost-effective solution to the problem. Deep learning (DL) models can work more efficiently and effectively than current climate model simulators that are highly computationally expensive. Our work will build on top of deep learning based predictive models where we will extend them from fixed treatment to continuous treatment setting. Table \ref{tab:tsci-methods} shows a holistic comparison of some of the time-series based causal inference methods and their capabilities to handles different causal inference scenarios.

\begin{table*}[!htbp]
\centering
\caption{Comparison of TCINet with existing time-series causal inference methods.}
\label{tab:tsci-methods}
\begin{tabular}{@{}cccccc@{}}
\toprule
Method &
  \begin{tabular}[c]{@{}c@{}}Binary/\\ fixed treatment\end{tabular} &
  \begin{tabular}[c]{@{}c@{}}Continuous \\ treatment\end{tabular} &
  \begin{tabular}[c]{@{}c@{}}Time varying \\ treatment\end{tabular} &
  \begin{tabular}[c]{@{}c@{}}Time varying\\  covariates\end{tabular} &
  \begin{tabular}[c]{@{}c@{}}Applicable on \\ Earth Science\end{tabular} \\ \midrule
Difference in Difference~\cite{did2011} &  \cmark   &    \xmark &  \xmark   & \xmark    &   \xmark  \\
Causal Impact~\cite{causalimpact2015} & \cmark  & \cmark & \cmark & \xmark &   \cmark  \\
CIV~\cite{CIV-2022}     &  \cmark   &   \cmark  &   \cmark  &   \cmark  &   \cmark  \\
CRN~\cite{bica2020estimating}  &   \cmark  &  \xmark   &   \cmark  &  \cmark   &   \xmark  \\
MSM~\cite{robins2000marginal} & \cmark & \xmark  & \cmark & \xmark  &   \xmark  \\
R-MSN~\cite{lim2018forecasting}  &   \cmark   &  \xmark    &  \cmark   &  \cmark   &   \xmark\\
Time-series Deconfounder~\cite{bica2020time} &  \cmark   &  \xmark    &  \cmark   &  \cmark   &   \xmark  \\
TCINet (ours)                  & \cmark & \cmark & \cmark & \cmark & \cmark \\ \bottomrule
\end{tabular}
\end{table*}

\section{Datasets}
To evaluate our model, we first generate synthetic data with time-delayed continuous treatment and time-varying covariates. We further provide details of the real world observational dataset pertaining to our research problem.
\subsection{Synthetic Data}
\label{sec:synthetic_data}
Using gaussian white noise, we generate four non-linear time-series given in Equations 4 to 7, mimicking the non-linearity in dynamic climate models. 

\begin{figure}[!htb]
  \centering
\includegraphics[width=0.6\linewidth]{ 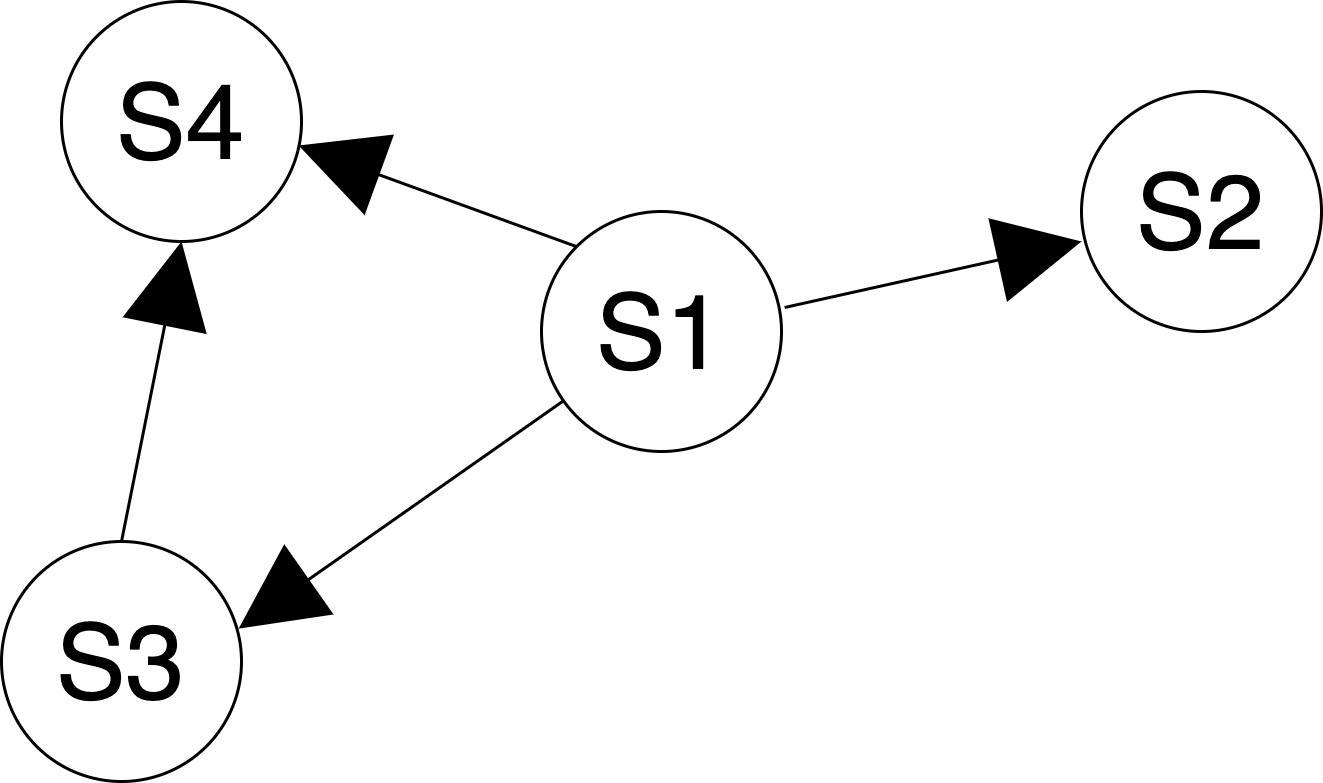}
  \caption{Causal graph of non-linear synthetic data.}\label{fig:graph-nonlinear}
\end{figure}

The corresponding true causal graph for these time-series is given in Figure \ref{fig:graph-nonlinear}. Here, we have taken $S3$ to be the treatment and $S4$ as the potential outcome. $S1$ and $S2$ will be considered as covariates where $S1$ is an observed time-varying confounder of both treatment and outcome. To generate counterfactuals, we intervene on $S3$, in two settings. First, we intervene on $S3$ as fixed treatment with binary values of $[0,1]$ to generate counterfactual values of $S4$. Next, we intervene on $S3$ by increasing $S3$ by $10\%$ and generate corresponding $S4$ counterfactuals under continuous treatment. 
    \begin{equation}
        S1_t = cos(\frac{t}{10}) + log(|S1_{t-6} - S1_{t-10}| + 1) + 0.1 \varepsilon 1 
    \end{equation}
    \begin{equation}
    S2_t = 1.2 e^{\frac{S1^2_{t-1}}{2}} + \varepsilon 2
    \end{equation}
    \begin{equation}
    S3_t = -1.05 e^{\frac{-S1^2_{t-1}}{2}} + \varepsilon 3
    \end{equation}
    \begin{equation}
    S4_t = -1.15 e^{\frac{-S1^2_{t-1}}{2}} + 1.35e^{\frac{-S3^2_{t-1}}{2}} + 0.28 e^{\frac{-S4^2_{t-1}}{2}} + \varepsilon4
    \end{equation}

\subsection{Observational Arctic Data}
We used observational sea-ice and reanalysis atmospheric and meteorological data which is available from 1979 till present. The reanalysis data is available with open access and can be obtained from European Centre for Medium-Range Weather Forecasts (ECMWF) ERA-5 global reanalysis product \cite{ERA5_URL}. Whereas the sea ice concentration (SIC) values are obtained from Nimbus-7 SSMR and DMSP SSM/I-SSMIS passive microwave data version 1 \cite{Cavalieri1996} provided by the National Snow and Ice Data Center (NSIDC).
The original data format is spatiotemporal from which we generated spatially averaged time-series combining sea ice extent values, oceanic and atmospheric variables. For this, daily gridded data over the regions of Barents Sea and Kara Sea, during 1979-2018, has been averaged using area-weighted method. The details of these variables are enlisted in Table \ref{tab:vartable}. 
\begin{table}[!htbp]
\caption{Variables in the Arctic Dataset}
\label{tab:vartable}
\begin{center}
\begin{small}
\begin{sc}
\begin{tabular}{cccc}
\toprule
\bfseries Variable & \bfseries Range & \bfseries Unit\\
\midrule
specific humidity & [0,0.1] & kg/kg\\
shortwave radiation & [0,1500] & $W/m^2$ \\
longwave radiation & [0,700] & $W/m^2$ \\
rainfall rate & [0,800] & mm/day\\
sea surface & & \\
temperature & [200,350] & K  \\
air temperature & [200,350] & K  \\
greenland & & \\
blocking index & [5000,5500] & m  \\
sea ice extent & [4, 13] & Million $Km^2$ \\
\bottomrule
\end{tabular}
\end{sc}
\end{small}
\end{center}
\vskip -0.1in
\end{table}

\section{Methodology}
\begin{figure*}[!htbp]
    \centering\includegraphics[width=0.8\textwidth]{ 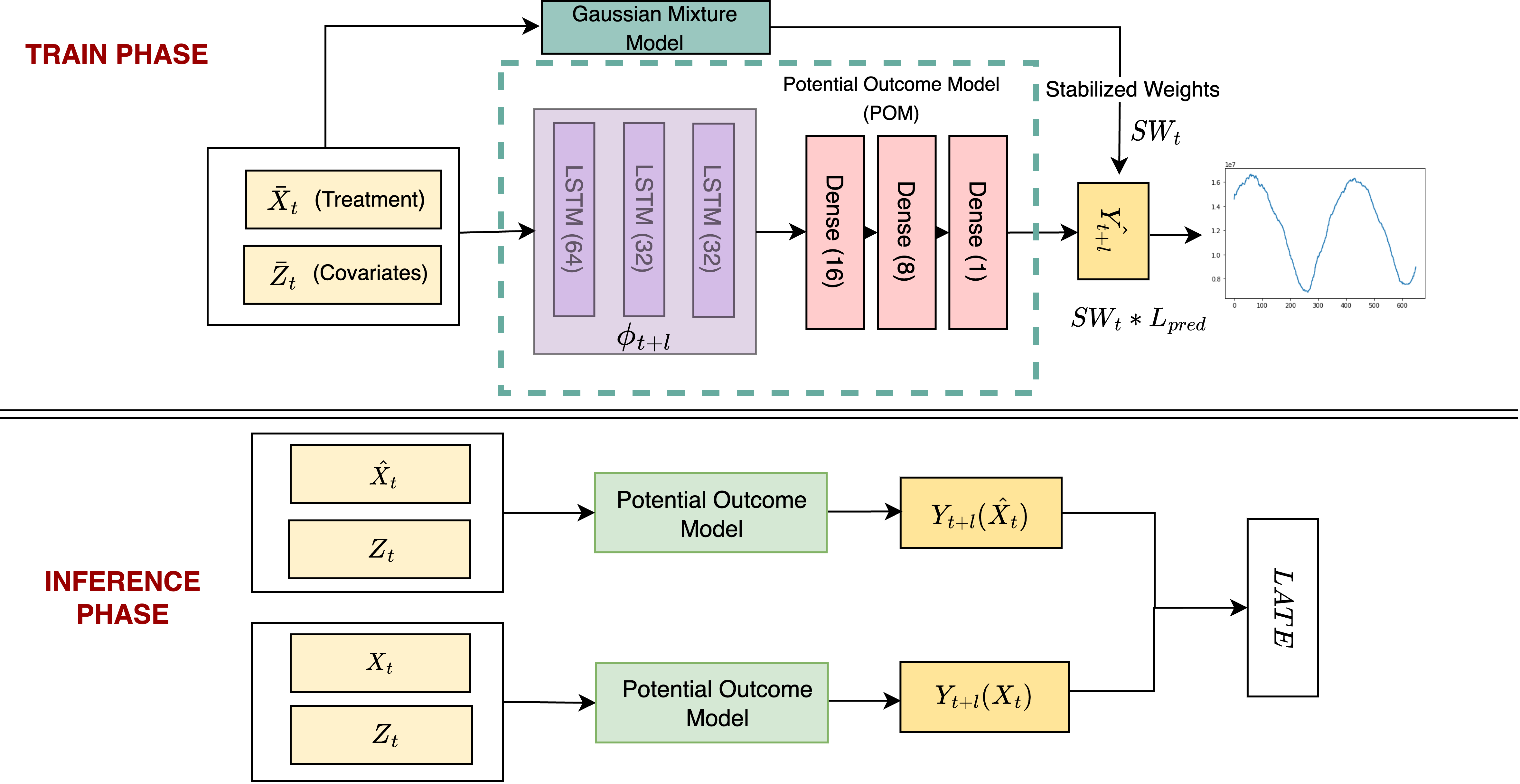}    
    \caption{Training and inference/test phase pipeline of our proposed TCINet model.}
    \label{fig:tci-pipeline}
\end{figure*}

Following the same principle of meta-learning used in existing deep learning based causal inference approaches, we propose a time-varying causal inference model, called Time-series Causal Inference Network (TCINet), on top of our previous work on LSTM based sea-ice forecasting model \cite{AliLSTMarXivNew}. The training and inference phases of our TCINet pipeline are illustrated in Figure \ref{fig:tci-pipeline}. In the training phase, time-delayed treatments $X_{t-l}$ and time-varying covariates $Z_{t-l}$ are fed to our potential outcome model (see Subsection~\ref{sec:pom}). To balance the bias due to time-varying covariates, we leverage Gaussian mixture modeling to compute stabilized weights 
(see Subsection~\ref{sec:balancing}). We also define a custom weighted loss to incorporate the balancing weights into our potential outcome model (see Subsection~\ref{sec:loss}). In the inference phase, we perturb the treatment variable and feed it to the pretained outcome model to make factual and counterfactual predictions (see Subsection~\ref{sec:inference}). We further explain how we estimate uncertainty during inference arguing on the feasibility of bootstrapping for time-series data.

\subsection{Balancing Time-varying Covariates}
\label{sec:balancing}
Balancing is a treatment adjustment strategy that aims to deconfound the treatment from outcome by forcing the treated and control covariate distributions as close as possible. When conducting observational studies, researchers often face the challenge that treatment assignment is not randomized, leading to potential confounding variables that can bias the treatment effect estimates. Inverse Probability of Treatment Weights (IPTW)~\cite{rubin1974estimating} is a statistical technique used in causal inference to address confounding bias in observational studies. IPTW generates a pseudo-population in which treatments are independent of confounders. To calculate IPTW, we first need the predicted probabilities of the observed treatments given the covariates. This is also known as the propensity score, given by $prob(X | Z = z)$.  When treatment and confounders are time-varying, these IPTW weights for time-fixed treatments need to be generalized. For a time-varying treatment $\bar{X_t} = (X_1, X_2,...,X_{t})$ and time-varying covariate $\bar{Z_t} = (Z_1, Z_2, ...,Z_{t})$, the IP weights for every timestep $t$ are given by \cite{huffman2018covariate}:
\begin{equation}
    IPTW(l) = \Pi_{t=1}^{l}\frac{1}{f(\bar{X_t}|\bar{Z_t})}
    \label{eq:iptw}
\end{equation}

Here, $l$ represents the lag or length of treatment sequence, $f(.)$ is the propensity score model, widely implemented using logistic regression following marginal structural modeling technique \cite{robins2000marginal}. However, the propensity scores that are near $0$ or $1$ can yield extreme IPTW weights, leading to unstable estimates and inflated variances. To tackle this, \cite{robins2000marginal} proposed the stabilized weights in which the IPTW is multiplied by the probability of receiving treatment, as given in Equation \ref{eq:sw}. Stabilized weights offer greater stability and reduce the variance in treatment effect estimation, which can improve the precision of the estimates. They are generally preferred in practice because of their improved numerical properties and stability. 

\begin{equation}
    SW(l) = \frac{\Pi_{t=1}^{l}f(X_t|\bar{X}_{t-1})}{\Pi_{t=1}^{l}f(X_t|\bar{X}_{t-1},\bar{Z}_{t})}
    \label{eq:sw}
\end{equation}

Here, $f(.)$ represents the probability density function (PDF) of treatment at every timestep given covariates and treatment history. In case of binary or discrete treatment, the PDF can be estimated using logistic regression or sigmoid function. However, in case of continuous treatment such as our case, this estimation becomes complex as it requires a parametric model to estimate the PDF at every stage $t$ \cite{huffman2018covariate}.

We implement a Gaussian Mixture Model (GMM) \cite{reynolds2009gaussian} to estimate the probability density of treatment $X_t$ at every timestep $t$. The step-by-step implementation of GMM for calculating stabilized weights is given in Algorithm \ref{algo:gmm_conditional_pdf}. We refer to the conditional probability densities in Equation \ref{eq:sw} as $X\_pdf$ and $XZ\_pdf$ in our algorithm. Whereas, the mean $\mu$, covariance $\Sigma$ and parameter $\alpha$, estimated as mixing co-efficients, are all learned by the GMM model. First, we fit the GMM model on treatment history and covariates to learn these parameters. We then estimate the probability density of $X_t$ given these parameter values at every timestep $t$ using Equation \ref{eq:pdf}.

\begin{multline}
    f(X_t | \mu, \Sigma) = \\ (\frac{1}{(2\pi)^{d/2}\sqrt{|\Sigma|}}) \exp\left[-\frac{1}{2}(X_t - \mu)^T  \Sigma^{-1} (X_t - \mu)\right]
    \label{eq:pdf}
    \end{multline}


\begin{algorithm}

  \caption{Stabilized Weights for Continuous Treatment}
  \label{algo:gmm_conditional_pdf}
      \KwData{Treatment  Data: $X$, Treatment History: $\bar{X}_{\text{hist}}$, Time-varying Covariates: $\bar{Z}$}
  \KwResult{Stabilized Weight Estimates $SW$}
    \SetKwFunction{FPdf}{PDF\_calc}
    \SetKwProg{Fn}{Function}{:}{}
    \Fn{\FPdf{$X$, $\bar{X}_{\text{hist}}$, $\bar{Z}=[ ]$}}{
    \tcp{Concatenate the treatment history and covariates}
      $\olsi{XZ} \leftarrow concat(\bar{X}_{\text{hist}}, \bar{Z}) $ \;
      $l \leftarrow$ length of sequence $XZ$\;
      \For{$i \leftarrow 1$ \KwTo $l$}{
      $n_{\text{comp}} \leftarrow$ Number of components for GMM\; 
    \tcp{Create a GMM object} 
    $gmm \leftarrow \text{GaussianMixture($n_{\text{comp}}$)}$ \; 
      \tcp{Fit the GMM model}
    $gmm$.fit($\olsi{XZ_i}$)\; 
      \tcp{Extract model parameters:}
    ($\alpha, \mu, \Sigma) \leftarrow (gmm$.weights, $gmm$.means, $gmm$.covariances)\;
    \tcp{Estimate PDF for every component}
    \For{$j \leftarrow 1$ \KwTo $n_{\text{comp}}$}{
      $pdf_{\text{comp}}[j] \leftarrow (\frac{1}{(2\pi)^{d/2}\sqrt{|\Sigma_j|}}) *  \exp\left[-\frac{1}{2}(X_i - \mu_j)^T  \Sigma^{-1}_j (X_i - \mu_j)\right]$\;}
    \tcp{Sum PDF over all components}
    $pdf[i] \leftarrow \sum_{j=1}^{n_{\text{comp}}} (pdf_{\text{comp}}[j] \times \alpha[j])$\;
    }
    \tcp{Take product of PDFs over all sub-sequences}
    $pdf_{\text{product}} \leftarrow$ $\Pi_{i=1}^{l} pdf[i]$
    \KwRet{ $pdf_{\text{product}}$}
    }
    $X\_pdf \leftarrow $PDF\_calc($X$, $\bar{X}_{\text{hist}})$ \;
    $XZ\_pdf \leftarrow $PDF\_calc($X$, $\bar{X}_{\text{hist}}$, $\bar{Z})$ \; 
    \tcp{Calculate stabilized weights at every timestep}
  \For{$k \leftarrow 1$ \KwTo $t_{\text{timesteps}}$}{
    $SW[k]$ $\leftarrow \frac{X\_pdf[k]}{XZ\_pdf[k]}$\;
    }

\end{algorithm}

\subsection{Potential Outcome Model (POM)}
\label{sec:pom}
We develop an LSTM-based prediction model as our potential outcome model (POM), following the S-learner approach \cite{koch2021deep}.
POM takes in input a 3D tensor of shape $N \times T  \times F$. Here $N$ represents the mini-batch size, $T$ represents the time lag and $F$ comprises the covariates and the treatment variable at timestep $t$. The model comprises three LSTM layers with RELU activation, where first two many-to-many (also called seq2seq) layers are followed by a Dropout layer to cater uncertainty estimation. These seq2seq layers take in a sequence of input of length $l$ and learn the latent representations $\phi$ of treatment and covariates. The third LSTM layer is a many-to-one layer succeeded by three fully connected Dense layers with linear activation. The purpose of these layers is to combine the learned representations to jointly predict the potential outcome $Y_{t+l}$ at timestep $t+l$ where $l$ is the time-dependency or lag. The model is compiled using Adam optimizer using the early stopping technique.

For a joint input, the model will learn mixed representations of covariates and treatment. This will be problematic in causal effect estimation as we want to keep the covariates independent of the intervention on the treatment variable. This is where the balancing strategy comes into play. To debias the confoundedness, we use Gaussian Mixture Model (GMM), as discussed in Subsection~\ref{sec:balancing}, to get the stabilized weights $SW_t$ for time-varying treatment at each timestep $t$ given the confounders. To train POM to make weighted predictions for potential outcome $Y_{t+l}$, we implement a custom-weighted loss function.

\subsection{Custom Weighted Loss}
\label{sec:loss}
We introduce the stabilized weights into the POM model using a custom weighted loss technique, to regularize the predictive model. The SW weights $SW_t$ calculated using GMM are inducted to the predictive model loss $L^{pred}$. This implies the final loss of the model will be a weighted average of prediction loss over observed data points $N$, as shown in Equation \ref{eq:causal-loss}, where $L^{pred}$ is the mean squared error (MSE) loss. 
\begin{equation}
    L_{tot} = \frac{1}{N}\sum_{t=1}^{N}SW_{t} *  L_t^{pred}
    \label{eq:causal-loss}
\end{equation}

\subsection{Inference}
\label{sec:inference}
Once the predictive model, namely POM, is trained on the observational data, the next step is to predict factual and counterfactual outcomes. We do this by perturbing the treatment variables at every timestep while retaining the observed values of time-varying covariates. The updated data is fed to the model to make counterfactual predictions while factual predictions are made without performing any nudging or intervention on the treatment variable. Once we have both predictions for all timesteps, we calculate LATE using Equation \ref{eq:late}.

To gain confidence in the predicted counterfactual values, we analyze the predictive skill of the underlying deep learning model and measure the model's epistemic uncertainty. Bootstrapping is a  common approach used for quantifying model uncertainty in causal inference techniques \cite{smith2022introduction}, however, bootstrapping will lead to two potential problems in case of our data. First, bootstrapping requires random sampling of data for train and test split but sampling randomly from time-series data will corrupt the sequential pattern and lead to unrealistic results. Second, bootstrapping involves retraining the model every time a random number of samples are taken from the data. In case of TCINet, it will be computationally expensive to retrain the deep learning model \textit{n} number of times as required by bootstrapping. 

We therefore take an alternative approach, where we train the TCINet modules POM and GMM once and make predictions \textit{n} times for each interventional scenario. We then calculate the mean and standard deviation of these predictions. The ATEs are recorded for observational data after making predictions for each case 50 times with a 95\% confidence interval. 

\section{Results \& Analysis}
In this section, we report our experimental setup and results obtained on synthetic and observational data, followed by a critical analysis of our findings using RMSE, LATE and PEHE scores.
\subsection{Evaluation Metrics}
\subsubsection{Root Mean Square Error (RMSE)}
We evaluate the performance of our predictive models using the Root Mean Square Error (RMSE) which can be only calculated for factual observational data but cannot be done for counterfactual predictions. 
\begin{equation}
    RMSE = \sqrt{\frac{1}{N} \Sigma_{i=1}^N(Y_i - \hat{Y_i})^2}
    \label{eq:rmse}
\end{equation}

\subsubsection{Precision in Estimated Heterogeneous Effects (PEHE)}
This metric is commonly used in machine learning literature for calculating the average error across the predicted ATEs \cite{hill2011bayesian}. PEHE metric, measuring causal effect estimation skill, can only be calculated for synthetic data which has ground truth information.
\begin{equation}
    \sqrt{PEHE} = \sqrt{\frac{1}{N} \Sigma_{i=1}^N(ATE_i - \hat{ATE_i})^2}
\end{equation}

\subsection{Experimental Setup}
We implement TCINet using Keras functional API with TensorFlow backend. The model has a total of $40,551$ trainable parameters. We compile the model using Adam optimizer with a $0.001$ learning rate and train it using early stopping technique. We train three variants of our model depending upon the underlying balancing strategies used in custom weighted loss: TCINet with SW weights using GMM which we refer to as TCINet-GMM, TCINet with IPTW weights using Logistic Regression model, which we refer to as TCINet-LR; and TCINet without any weighting using standard Mean Squared Error loss which we refer to as TCINet$^-$ in Table~\ref{tab:res-model-fix}.

\subsection{Results on Synthetic Data}
We report our results on the three variants of the model; TCINet-GMM, TCINet-LR and TCINet and compare them with two state-of-the-art (SOTA) methods: time-varying CIV technique~\cite{CIV-2022} and time-invariant Causal Impact~\cite{causalimpact2015} causal inference methods. We evaluate both the CIV and Causal Impact method using the synthetic dataset to measure the causal effect from the cause $S3$ to the target variable $S4$ in case of the fixed and continuous treatments explained in \ref{sec:synthetic_data}. We report these results in Table \ref{tab:res-model-fix}. 
Comparing the performance of three TCINet variants in Table \ref{tab:res-model-fix}, we notice that all variants have marginal differences in RMSE scores, however, we see substantial differences in ATE estimation by these models. This performance difference is also evident from the low PEHE scores for TCINet-GMM in Table \ref{tab:res-model-fix}.  Since CIV does not provide RMSE values on factual estimation, we compare its performance based on estimated ATE values. Though CIV is an easier model to implement, we notice that in both cases, i.e., fixed and continuous treatment effect estimation, CIV performs poorly as compared to TCINet variants and Causal Impact, which gives us more confidence in our model performance. It is important to note here that Causal Impact provides the second best performance in case of fixed and continuous treatments, however, inherently Causal Impact cannot work with time-varying covariates and is therefore not suitable for our case. Moving forward, we analyze the observational data using TCINet-GMM owing to its superior performance.

\begin{table}[!htbp]
\caption{Causal Inference Models Performance on Synthetic Data under Fixed and Continuous Treatments for one-step ahead Prediction (True ATE = -0.0514)}
\label{tab:res-model-fix}
\begin{center}
\begin{small}
\begin{sc}
\begin{tabular}{lllll}
      \toprule 
      \bfseries Model  & \bfseries Test & \bfseries Estimated & \bfseries PEHE \\
        & \bfseries RMSE & \bfseries LATE & \\
      \midrule 
      \multicolumn{4}{c}{Fixed Treatment}      \\ \hline
      \textbf{TCINet$^-$} & 1.079 & -0.040 & 1.132\\
      \textbf{TCINet-LR} & 1.142 & -0.037 & 1.227\\
      \textbf{TCINet-GMM} & 1.023 & -0.051 & 1.004 \\
      \textbf{CIV~\cite{CIV-2022}} & n/a & -0.219 & n/a \\
      \textbf{Causal Impact~\cite{causalimpact2015}} & n/a & -0.060 & 1.110  \\ \hline
      \multicolumn{4}{c}{Continuous Treatment} \\ \hline
      \textbf{TCINet$^-$} & 1.026 & -0.036 & 1.221\\
      \textbf{TCINet-LR} &  1.000  &  -0.049 & 1.143\\
      \textbf{TCINet-GMM} & 0.998 & -0.050 &  1.102\\
      \textbf{CIV~\cite{CIV-2022}} & n/a & 0.515 & n/a \\
      \textbf{Causal Impact~\cite{causalimpact2015}} & n/a & -0.040 & 1.112  \\
        \bottomrule 
    \end{tabular}
\end{sc}
\end{small}
\end{center}
\vskip -0.1in
\end{table}

\subsection{TCINet for Arctic Amplification}

After gaining confidence in the predictive skill of TCINet for synthetic data, we use the model to answer an important domain science question on the observational data as identified by Atmospheric scientists 
 \cite{huang2021summertime}: \textit{How does increased Greenland Blocking (GBI) affect summertime regional Arctic sea ice melting given snowfall rate and solar radiation data?}

The Greenland block is a ridge of high pressure that sits near or over Greenland. It is the normalised area-weighted 500 hPa geopotential height over the region $60 - 80^{\circ} N, 20 - 80^{\circ} W$. To identify the regions of interest and time lag by which GBI affects sea ice extent, we performed lagged correlation test between daily GBI values and regional sea ice extent given by \cite{Meier2023} for sixteen sub-regions. We conducted experiments for a lag of 0 to 30 days and found the highest correlation at day 8 between GBI and SIE in Barents Sea and Kara Sea (combined as BarKara Sea in our analysis).  

To answer the domain science question, we first trained TCINet-GMM on forty years of our observational data. We then predict sea ice extent by perturbing the values of summertime (June, July, August) GBI to the following four values: 1) 40-year-averaged daily GBI, 2) double GBI annual trend, 3) triple GBI annual trend, 4) quadruple GBI annual trend.

In our efforts to quantify the effects of increasing GBI on declining sea ice, we first make predictions for summertime (June, July, August) sea ice for mean daily GBI values. We then perturb the GBI values by increasing them by a multiplicative factor of the daily recorded trend, i.e. 0.039. Our interpretation of ATE in case of observational sea ice data is that it reflects the average increase or decrease in sea ice extent under interventional treatment. 

\begin{figure}[!htbp]
    \centering
    \includegraphics[width=0.5\textwidth]{ 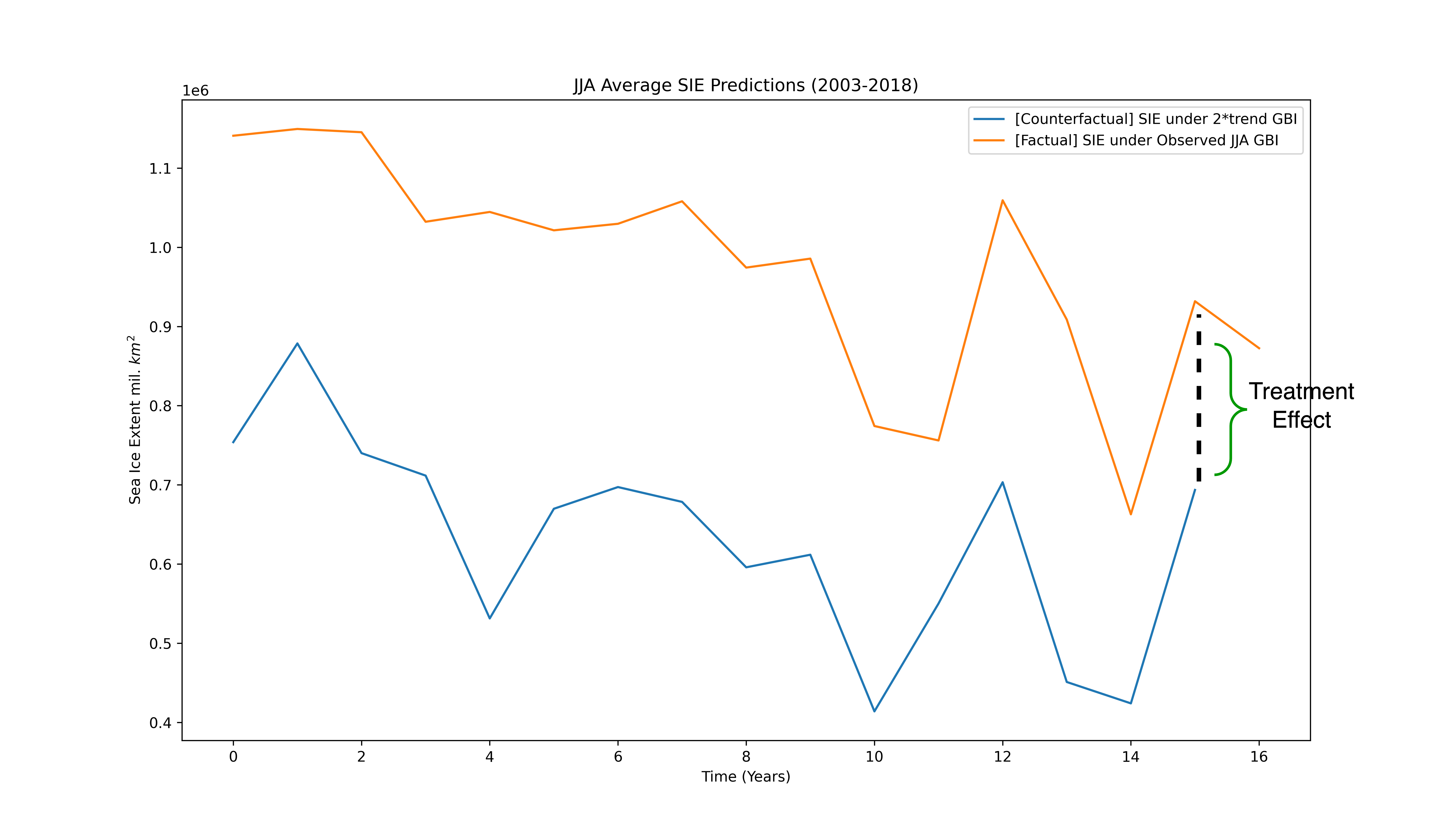}
    \caption{Comparison of annual mean sea ice extent (SIE) predictions given observational data (factual) versus predictions under interventional GBI (counterfactual) between 2003-2018. Here, each data point represents summer (JJA) mean SIE predicted for that year.}
    \label{fig:sie_under_gbi}
\end{figure}

As shown in Figure \ref{fig:sie_under_gbi}, we notice that increasing GBI leads to decrease in sea ice extent (blue line with counterfactual predictions). Quantitatively, our model predicts that the average daily sea ice extent value in JJA summer months would have decreased by 0.64, 0.65 and 0.69 million $km^2$ between 2003 to 2018, given the GBI was increased by 2, 3 and 4 times the daily trend. This aligns with the findings of \cite{huang2021summertime} where summertime low clouds play an important role in driving sea ice melt by amplifying the adiabatic warming induced by a stronger anticyclonic circulation aloft.  

\section{Discussion \& Future Work}
In this paper, we propose a deep learning based time-series inference method for time-varying causal inference under continuous treatment effects using stabilized weights. We introduce a probabilistic method of implementing stabilized weights through gaussian modeling. Through ablative study, we show how our proposed model balances confoundedness in case of time-delayed treatment. We presented one use-case of analyzing the causal relation between Greenland blocking and sea ice melt. Through experiments, we noticed our data-driven findings align with the literature on "increasing GBI leads to decreasing SIE". For our ongoing research, we will continue to analyze similar other use cases in the realm of Arctic Amplification, such as the effects of atmospheric processes on Arctic sea ice melt. We will further extend our work to spatiotemporal causal inference to explore the potential of neural networks in learning and answering important Earth Science questions in the presence of temporal and spatial confounders.

\section*{Acknowledgement}
This work is supported by NSF grants: CAREER: Big Data Climate Causality (OAC-1942714) and HDR Institute: HARP - Harnessing Data and Model Revolution in the Polar Regions (OAC-2118285). 

\bibliographystyle{abbrv}
\bibliography{ieee_ref}

\end{document}